\documentclass[10pt,twocolumn,letterpaper]{article}

\usepackage{btas}
\usepackage{times}
\usepackage{epsfig}
\usepackage{graphicx}
\usepackage{amsmath}
\usepackage{amssymb}
\usepackage{balance}
\usepackage{multirow}
\usepackage{soul}
\usepackage{booktabs,caption}
\usepackage[flushleft]{threeparttable}
\usepackage{placeins}
\usepackage{siunitx}
\usepackage[pagebackref=false,breaklinks=true,letterpaper=true,colorlinks,bookmarks=false]{hyperref}
\usepackage{fancyhdr}
\usepackage[skip=2pt]{caption} 
\usepackage{enumitem}
\usepackage{xcolor} 
\btasfinalcopy

\begin{document}

\title{Gender Privacy: An Ensemble of Semi Adversarial Networks for Confounding Arbitrary Gender Classifiers }

\author{Vahid Mirjalili\textsuperscript{ 1}  \qquad \qquad Sebastian Raschka\textsuperscript{ 2} \qquad \qquad \qquad Arun Ross\textsuperscript{ 1} \\ \;\;\;\;\;\; {\tt\small mirjalil@cse.msu.edu} \;\;\;\; \; {\tt\small mail@sebastianraschka.com}  \;\;\;  \;\;\;   {\tt\small rossarun@cse.msu.edu}  \\ 
\textsuperscript{1}  Computer Science \& Engineering, Michigan State University, East Lansing, USA \\
\textsuperscript{2} Department of Statistics, University of Wisconsin -- Madison, USA}


\maketitle
\fancypagestyle{plain}{
	\fancyhf{} 
	\fancyhead[C]{\textcolor{red}{Published in Proc. of IEEE 9th International Conference on Biometrics: Theory, Applications and Systems (BTAS), (Los Angeles, CA), October 2018.}}
	\fancyfoot[R]{}
	\renewcommand{\headrulewidth}{0pt}
	\renewcommand{\footrulewidth}{0pt}
}

\begin{abstract}

Recent research has proposed the use of Semi Adversarial Networks (SAN) for imparting privacy to face images. SANs are convolutional autoencoders that perturb face images such that the perturbed images cannot be reliably used by an attribute classifier (e.g., a gender classifier) but can still be used by a face matcher for matching purposes. However, the generalizability of SANs across multiple arbitrary gender classifiers has not been demonstrated in the literature. 
In this work, we tackle the generalization issue by designing an ensemble SAN model that generates a diverse set of perturbed outputs for a given input face image. 
This is accomplished by enforcing diversity among the individual models in the ensemble through the use of different data augmentation techniques. 
The goal is to ensure that at least one of the perturbed output faces will confound an arbitrary, previously unseen gender classifier. 
Extensive experiments using different unseen gender classifiers and face matchers are performed to demonstrate the efficacy of the proposed paradigm in imparting gender privacy to face images.
\end{abstract}

\section{Introduction}

\begin{quote}
``If this is the age of information, then privacy is the issue of our times.''~\cite{acquisti_privacy_2015}
\end{quote} 

Recent data breaches have not only caused tremendous financial and societal consequences~\cite{facebook_scandal_2018},
but have raised several privacy issues, urging a new body of legislation to protect users' privacy in e-corporates, governments, and international affairs~\cite{acquisti_predicting_2009}. In addition, the newly implemented General Data Protection Regulation
(GDPR)~\cite{eu_gdpr_2016} requires businesses to explicitly inform the users about the purpose of data collection, and prohibits any data processing beyond the stated purpose. 
In principle, privacy laws should grant users the right to determine for themselves which information about them to reveal and which to conceal~\cite{kindt_privacy_2016,acquisti_what_2013,acquisti_economics_2016,jourabloo_attribute_2015}. This has direct implications in the field of biometrics, which is the science of recognizing individuals based on their physical or behavioral traits.   
A person's biometric data, such as face or fingerprint images, may contain private information about the individual~\cite{jain_introduction_2011}. For example, soft biometric attributes such as age, gender, and race can be gleaned from face images~\cite{dantcheva_what_2016}. Recent advancements in machine learning~\cite{raschka_python_2017} have made it possible to automate the extraction of such information from biometric data of individuals stored in central databases~\cite{dantcheva_what_2016}. Extracting such information without an individual's consent can be considered a violation of their privacy~\cite{kindt_privacy_2016,acquisti_what_2013,acquisti_economics_2016,mirjalili_soft_2017}. In this regard, recent research has explored the possibility of modifying face images such that certain soft biometric attributes cannot be extracted, while the modified face image can still be used for biometric recognition purposes. 

\section{Related Work}

Extracting soft biometric attributes, such as age and gender from face images, has been extensively studied~\cite{becerra_age_2017,gunther_affact_2016,dantcheva_what_2016}. A wide range of methods have been employed, including those based on custom feature extraction techniques~\cite{bukar_automatic_2016} and those based on deep learning techniques~\cite{levi_age_2015,mansanet_local_2016,jia_gender_2016,castrillon_descriptors_2017,gunther_affact_2016}.  However,  imparting soft biometric privacy by confounding such attributes is a relatively recent research area. One of the earliest attempts goes back to 2014, when Othman and Ross~\cite{othman_privacy_2014} developed a method for modifying the gender attribute of face images while preserving their face matching utility. In 2015, Sim and Zhang~\cite{sim_controllable_2015} proposed a face de-identification approach where certain attributes were  preserved while others could be selectively perturbed.

Inspired by the notion of adversarial examples~\cite{szegedy_intriguing_2013,akhtar_threat_2018} and the study of robustness of facial attribute classifiers to adversarial examples~\cite{rozsa_are_2016_long}, Mirjalili and Ross~\cite{mirjalili_soft_2017} investigated the possibility of utilizing adversarial images for imparting gender privacy. The researchers were able to generate image perturbations targeting a specific gender classifier and showed that these perturbations could confound the gender classifier, while preserving the performance of a commercial face-matcher~\cite{mirjalili_soft_2017}.  


Although perturbed adversarial images have shown to be effective in confounding a {\em particular} classifier, the issue that these images may not adversarially affect other {\em unseen} classifiers limits their effectiveness in practical privacy applications.\footnote{The term ``unseen" indicates that the classifier or matcher was not used during the training stage.}  Adversarial images generated for a particular gender classifier may not generalize to another. Furthermore, in a real-world application, the knowledge of a gender classifier may not be available in advance; as a result, generating adversarial images for an unseen gender classifier would be difficult. 
To address this issue, Mirjalili \etal~\cite{mirjalili_semi_2018} developed an autoencoder called Semi Adversarial Network (SAN) for generating perturbed face images that could potentially generalize across unseen gender classifiers. They trained the SAN model using an auxiliary gender classifier and an auxiliary face matcher and evaluated the success of their model in producing output images that could confound two unseen gender classifiers, while preserving the recognition accuracy of an unseen face matcher. Although the accuracy of the two unseen gender classifiers were indeed confounded, yet, generalizability to a large number of unseen gender classifiers remains an open problem (see Section~\ref{sec:unseen-gender-classifiers}). Furthermore, a human observer may be able to correctly classify the gender of the perturbed images generated by their model (see Figure~\ref{fig:example-orig-ensans}), which means that, in principle, there exists an unseen gender classifier that can correctly recognize the gender of the perturbed images. In this paper, we formulate an {\em ensemble} technique to address the limitations of the previous SAN model and facilitate its generalizability to a large number of unseen gender classifiers.\footnote{The acronym SAN was simultaneously coined by two independent research groups. Cao \etal~\cite{cao_partial_2017} defined Selective Adversarial Networks for partial style transfer, and Mirjalili \etal~\cite{mirjalili_semi_2018} defined Semi Adversarial Networks for imparting privacy to face images. Here, we use SAN to refer to the latter.} In the context of this work, the \textit{generalizability} of a SAN model is defined as its ability to perturb face images in such a way that an {\em arbitrary} unseen gender classifier is confounded while an arbitrary unseen face matcher retains its utility.


The major contributions of this work are as follows:
\begin{itemize}[noitemsep,nolistsep]
\item Designing an ensemble of SANs to address the problem of generalizability across unseen gender classifiers.
\item Conducting large-scale experiments that convey the practicality and efficacy of the proposed approach. 
\item Ensuring that race and age attributes are retained in the perturbed face images.
\end{itemize}

\section{Proposed Method}
\begin{figure}
\begin{center}
   \includegraphics[width=1.05\linewidth]{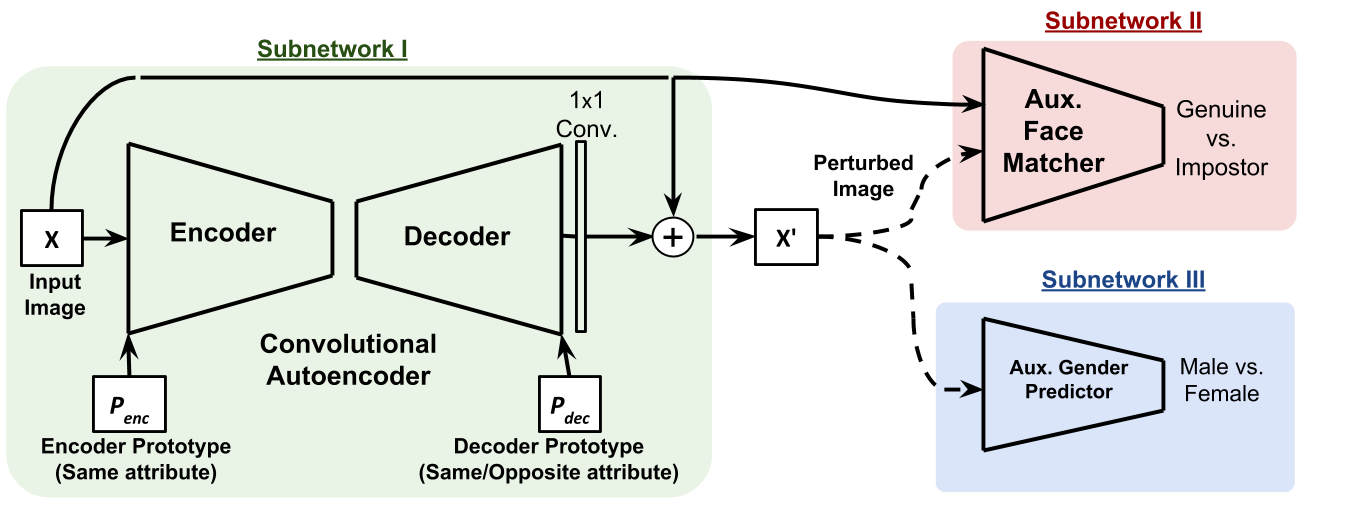}
\end{center}
   \caption{Architecture of the original SAN model~\cite{mirjalili_semi_2018}.}
\label{fig:san-arch}
\end{figure}

{\bf Previous SAN model}~\cite{mirjalili_semi_2018}:  The overall architecture of the individual SAN models  in the ensemble is similar to the SAN model proposed in~\cite{mirjalili_semi_2018} as shown in Figure~\ref{fig:san-arch}, but with a few modifications. The SAN model consists of a {\bf convolutional autoencoder} to perturb face images, a convolutional neural network (CNN) as an {\bf auxiliary face matcher}, and a CNN as an {\bf auxiliary gender classifier}. The pre-trained, publicly available VGG-face CNN~\cite{parkhi_deep_2015} is used as the auxiliary face matcher. The input gray-scale image is first fused with a {\bf face prototype} belonging to the same gender as the input image. Then $128$ feature maps are obtained from the last layer of the decoder, which are combined with the face prototype of the opposite gender using $1\times 1$ convolutions. The final image is then passed to both the auxiliary face matcher and the auxiliary gender classifier to compute its match score with the original input and its gender probability, respectively. During training, each input image is reconstructed by the autoencoder using both same-gender and opposite-gender prototypes to obtain two different outputs. Then, three different cost functions are used based on these outputs. First, a pixel-wise similarity measure between the input and the output from the same-gender prototype is used as a cost function to ensure that the autoencoder is able to construct realistic images. The second cost function is the $L^2$ distance between the face vector of the input image and those of the outputs to make the autoencoder learn to perturb face images such that the accuracy of the face matcher is  retained. The third cost term is the cross-entropy loss applied to the gender probabilities of the two outputs as computed by the auxiliary gender classifier, where the ground-truth label of the input image is used for the output of the same-gender prototype but the reverse is used for the output of the opposite-gender prototype.
\label{sec:ensemble-san-formulation}
\subsection{Ensemble SAN Formulation}

We assume that there exists a large set of gender classifiers $\mathcal{G}=\{G^1, G^2, ..., G^n\}$, where each $G^i(X)$ predicts the gender of a person based on a 2D face image, $X$. Furthermore, we assume a set of face-matchers denoted by $\mathcal{M}=\{M^1, M^2, ..., M^m\}$, where each $M^i(X_a, X_b)$ computes the match score between a pair of face images, $X_a$ and $X_b$. 
The goal of the work is to design an {\em ensemble} of $t$ SAN models, $\mathcal{S} = \{S^1, S^2, ..., S^t\}$, that can be shown to generalize to arbitrary gender classifiers. In particular, we demonstrate that for each face image $X$, $\mathcal{S}$ produces a set of outputs $\mathcal{S}(X) = \{Y^1, Y^2, ..., Y^t\}$ such that for each $G^i\in \mathcal{G}$, there exists at least one output $Y^j = S^j(X)$ that is able to confound $G^i$. At the same time, the outputs, $\mathcal{S}(X)$, can be successfully used for face recognition by the matchers in $\mathcal{M}$. 

\subsection{Diversity in Autoencoder Ensembles}

One of the key aspects of neural network ensembles is diversity among the individual network models~\cite{hansen_neural_1990}. Several techniques have been proposed in the literature for enhancing diversity among individual networks in an ensemble, such as seeding the networks with different random weights, choosing different network architectures, or using bootstrap samples of the training data~\cite{strauss_ensemble_2017,dietterich_ensemble_2000}.

In the context of SAN models, autoencoder diversity can be imposed in two ways: (a) through training on different datasets, and (b) by utilizing different auxiliary gender classifiers. 
Intuitively, an ensemble of classifiers can only be useful if individual classifiers do {\bf not} make similar errors on the test data~\cite{strauss_ensemble_2017, hansen_neural_1990, kuncheva_combining_2004}.
To benefit from ensembles, it is thus critical to ensure error diversity, which can be accomplished by assembling the ensemble from a diverse set of classifiers. A number of approaches to explicitly measure ensemble diversity have been reported in the literature \cite{kuncheva_combining_2004}.

\begin{figure}
\begin{center}
   \includegraphics[width=0.8\linewidth]{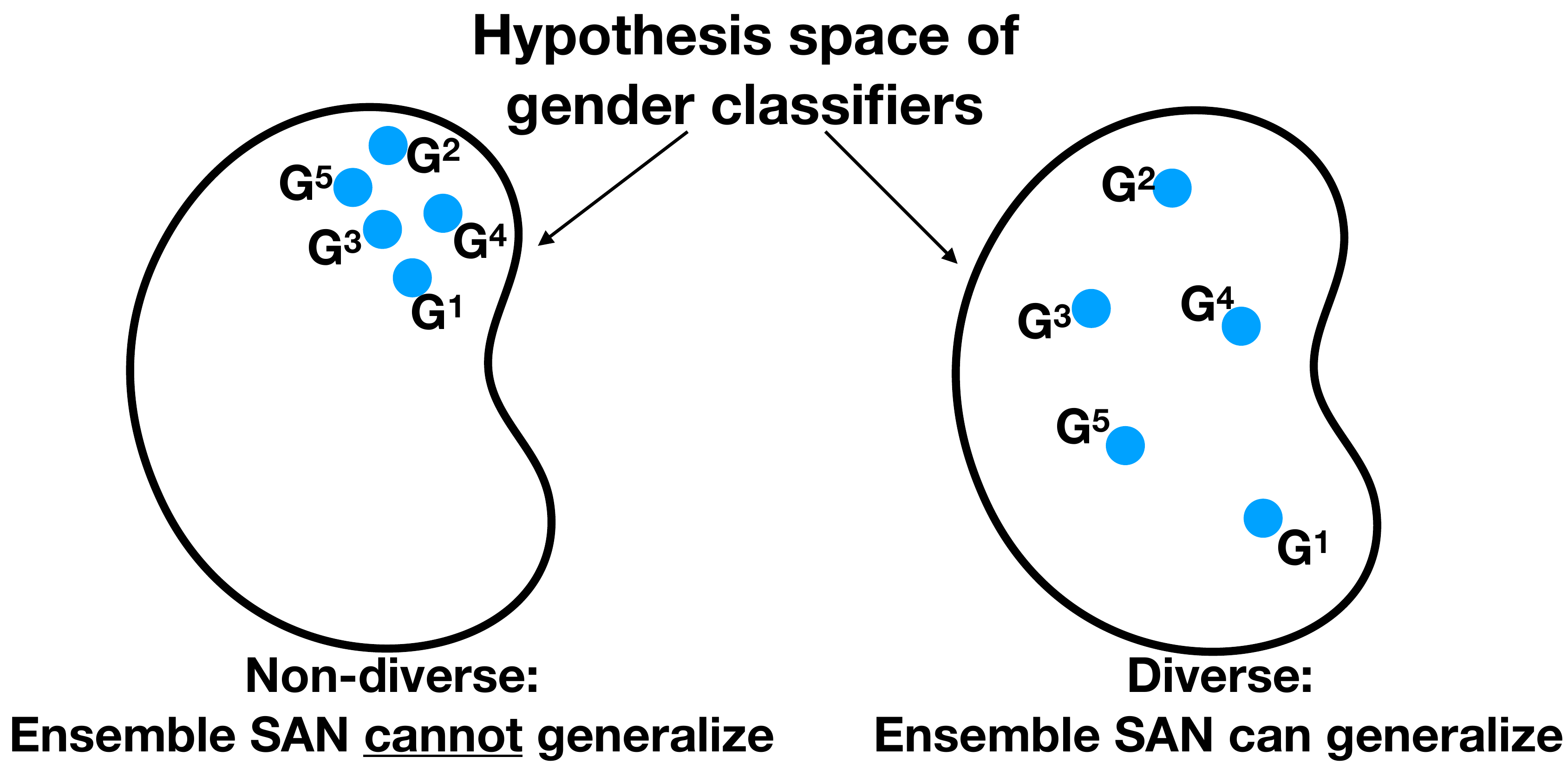}
\end{center}
   \caption{Diversity in an ensemble SAN can be enhanced through its auxiliary gender classifiers (see Figure~\ref{fig:san-arch}). When the auxiliary gender classifiers lack diversity, ensemble SAN cannot generalize well to arbitrary gender classifiers.}
\label{fig:diversity}
\end{figure}

Among the novel contributions of this work is the development of ensemble methods for SANs using oversampling and data augmentation techniques. As shown in Figure~\ref{fig:diversity}, if auxiliary gender classifiers that are used to build a SAN lack diversity, the ensemble SAN {\em cannot} generalize to arbitrary classifiers. Therefore, in order to ensure generalizability, we (1) diversify the auxiliary gender classifiers and (2) diversify the autoencoder component of the SANs during the training phase.

\subsection{Ensemble SAN Architecture}

The original SAN model used single-attribute prototype images, which were computed by averaging over all male and female images, respectively, in the training dataset~\cite{mirjalili_semi_2018}. 
However, this approach does not take other soft-biometric attributes into account, such as race and age, which increases the risk of introducing a systematic bias to the perturbed images if certain attributes are over- or under-represented in the training dataset. This issue is addressed in the current work. 

\begin{figure}[t!]
\begin{center}
   \includegraphics[width=0.9\linewidth]{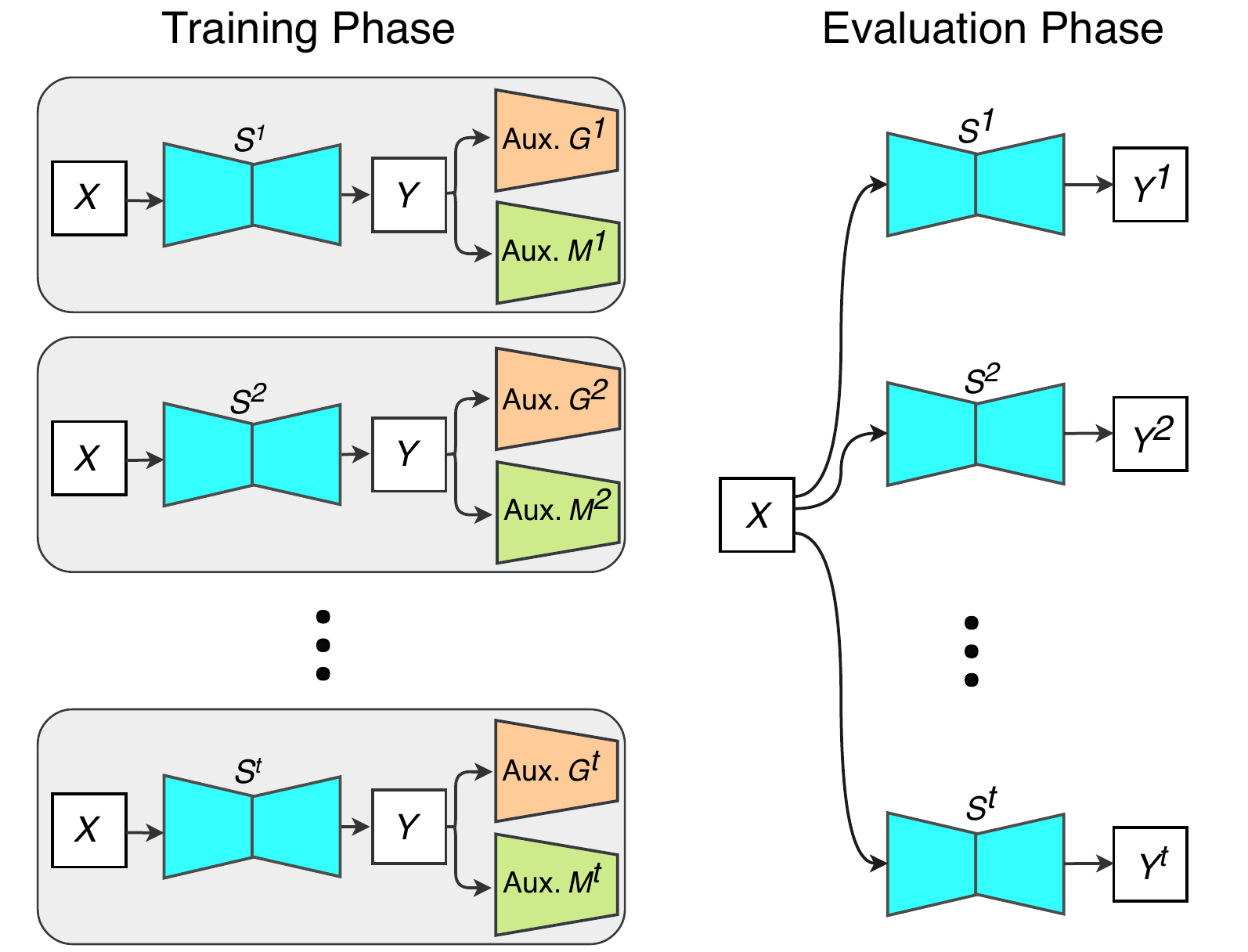}
\end{center}
   \caption{Schematic of the proposed ensemble of $t$ SAN models. During the training phase, each SAN model, $S^i$, is associated with an auxiliary gender classifier $G^i$ and an auxiliary face matcher $M^i$ (common across all SANs). During the evaluation phase, the trained SAN models are used to generate $t$ outputs $\{Y^1, Y^2, ..., Y^t\}$.}
\label{fig:ensan-diagram}
\end{figure}

{\bf Proposed Ensemble Model:}  The overall architecture of the proposed model is shown in Figure~\ref{fig:ensan-diagram}. The ensemble consists of $t$ individual SAN models that are trained independently as will be discussed later. Each model is associated with an individually pre-trained auxiliary gender classifier and a pre-trained auxiliary face matcher.\footnote{The term auxiliary is used to indicate that these gender classifiers and face matchers are {\em only used during training} and {\em not} associated with any of the ``unseen" gender classifiers and face matchers that will be used in the evaluation phase.}  After the training of a SAN model has been completed, the auxiliary networks (gender classifier and face matcher) are discarded, and each SAN model $S^j$ is used to generate an output image $Y^j$ ($j \in \{1, ..., t\}$) from an input image $X$, which results in a total of $t$ output images.

\begin{figure}
\begin{center}
   \includegraphics[width=0.7\linewidth]{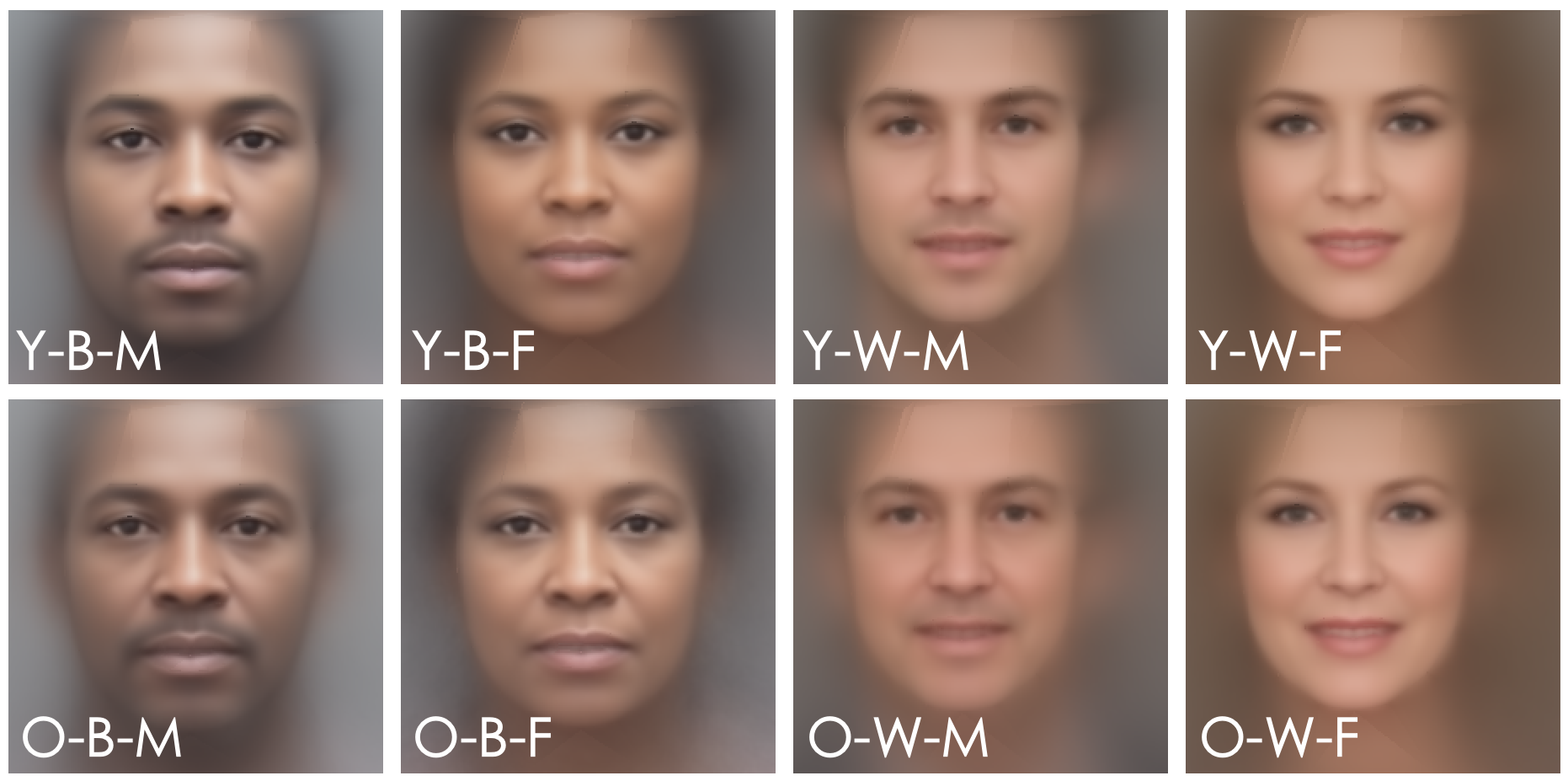}
\end{center}
   \caption{Face prototypes computed for each group of  attribute labels. The abbreviations at the bottom of each image refer to the prototype attribute-classes, where Y=young, O=old, M=male, F=female, W=white, B=black.}
\label{fig:gender-prototypes}
\end{figure}

We further propose that taking attributes other than just the attribute of interest (i.e., gender) into account reduces side-effects such as modifications to the race and age of an input image. Considering three binary attributes, gender (male, female), age (young, old), and race (black, white), we can categorize an input image into one of eight disjoint groups. For each group, we generate a prototype image, which is the average of all face images from the training dataset that belong to that group. Hence, given eight distinct categories or groups, eight different prototypes are computed. Next, an opposite-attribute prototype is defined by flipping one of the binary attribute labels of an input image. For example, if the input image had the attribute labels \{young, female, white\}, the opposite-gender prototype chosen for gender perturbation would be \{young, male, white\}. The face prototype for each group is shown in Figure~\ref{fig:gender-prototypes}, and is computed by aligning the corresponding faces onto the the average face shape of each group.

The similarities and differences between the originally proposed SAN model and the ensemble SANs developed in this work are summarized below:
\begin{itemize}[noitemsep]
\item The autoencoder, auxiliary gender classifier, and auxiliary face matcher architectures are similar to the original SAN model.
\item In contrast to the original SAN model, we construct face image prototypes to reduce alterations to non-target attributes such as age and race.
\item Instead of training a single SAN model, we create an ensemble of diverse SAN models that extend the range of arbitrary gender classifiers that can be confounded while preserving the utility of arbitrary face matchers.
\end{itemize}

\subsection{Ensemble of SANs: Training Approach}

\begin{figure}
\begin{center}
   \includegraphics[width=1.01\linewidth]{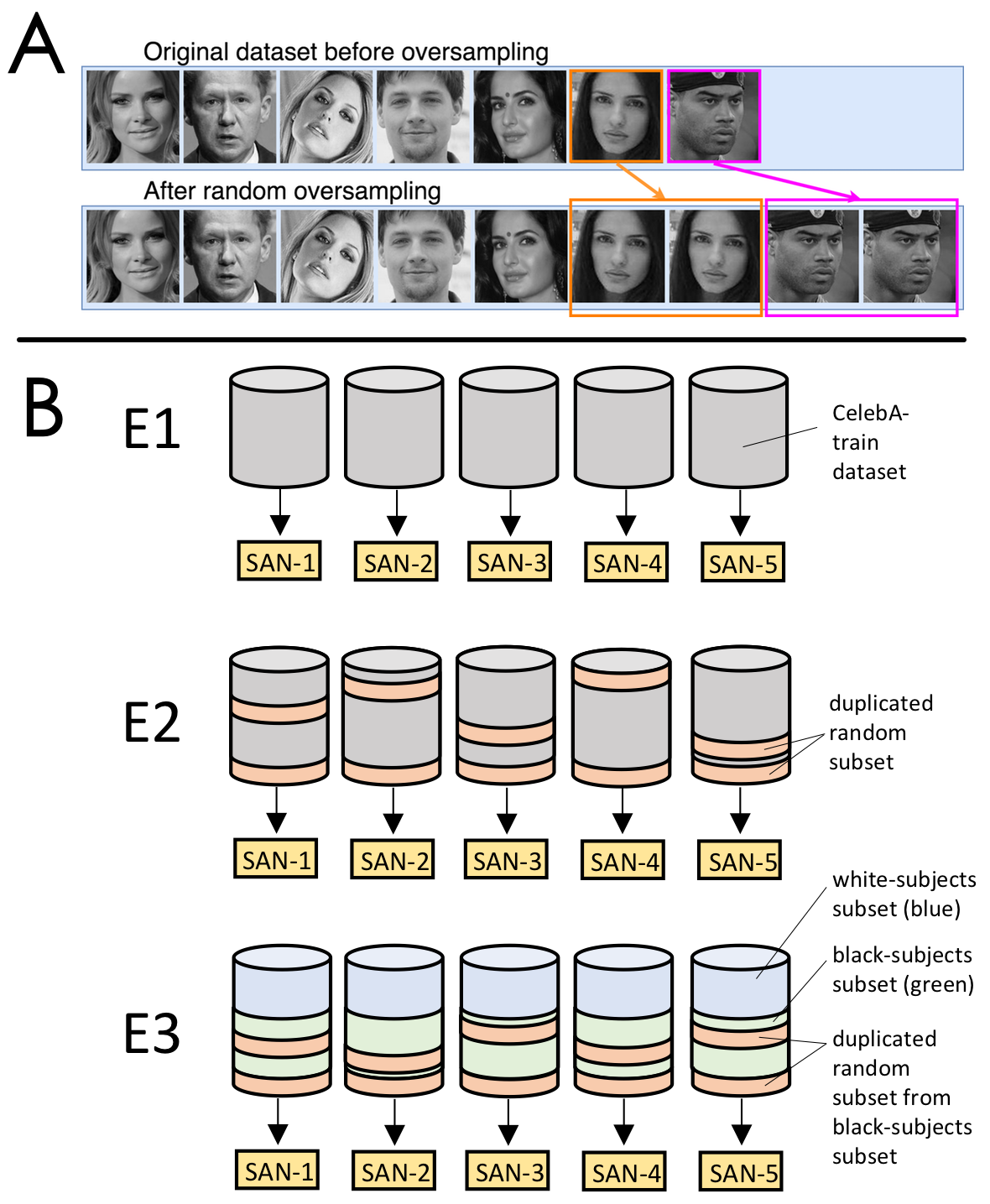}
\end{center}
   \caption{An example illustrating the oversampling technique used for enforcing diversity among SAN models in an ensemble. A: A random subset of samples are duplicated. B: Different Ensemble SANs (E1, E2, and E3) are trained on the CelebA-train dataset. SANs of the E1 ensemble are trained on the same dataset with different random seeds. In addition to using different random seeds, E2 SAN models are trained on datasets created by resampling the original dataset (duplicating a random subset of the images). Finally, for E3, a random subset of black subjects was duplicated for training the different SANs in the ensemble.}
\label{fig:example-oversampling}
\end{figure}

To obtain a diverse set of SAN models, we trained the individual SAN models using different initial random weights. Further, we enhanced the diversity among the models by designing three different {\em training schemes} for the auxiliary gender classifier component of the SAN model as illustrated in Figure~\ref{fig:example-oversampling} and further described below.

\begin{itemize}[noitemsep]
\item {\bf E1 (regular):} Consists of five SANs, where the auxiliary gender classifier in each SAN model was initialized with different initial random weights. The models were trained on the CelebA training partition without resampling.
\item {\bf E2 (subject-based oversampling):} Consists of five  SANs similar to E1, but in addition to choosing different initial random weights for the auxiliary gender classifiers, we applied a resampling technique by  duplicating each sample from a random subset of subjects (representing 10\% of the images in the training set).
The selected subjects are disjoint across the five models, and the samples are duplicated four times.
\item {\bf E3 (race-based oversampling):} Five SANs were trained, similar to E1 and E2, but instead of resampling a random subset of subjects as in E2, we resampled instances of the minority race represented in the CelebA dataset to balance the racial distribution in the training data. In particular, a random $10\%$-subset of black samples was duplicated $40$ times, that is, $10\%$ of the black samples were copied $40$ times and appended to the training dataset.
\end{itemize}

\subsection{Datasets}

We used five face image datasets in this work, viz., CelebA~\cite{liu_deep_2015_long}, MORPH~\cite{ricanek_morph_2006}, LFW~\cite{huang_labeled_2007_long}, MUCT~\cite{milborrow_muct_2010}, and RaFD~\cite{langner_presentation_2010}. The details of the datasets, and how they were used in this work, are summarized in Table~\ref{tab:datasets}. Furthermore, the CelebA and MORPH datasets were split into non-overlapping training and test partitions, such that the train and test partitions are subject-disjoint (i.e., if a dataset contained multiple poses of the same person, these were all included either in the training set or the test set but not both). CelebA-train was used for training the auxiliary gender classifiers under the three schemes mentioned in the previous section, as well as for training all the individual SAN models. The face prototypes were computed using the CelebA-train and MORPH-train datasets. The remaining datasets were used for evaluating the performance of the SAN models on unseen gender classifiers and unseen face matchers.

\begin{table}[h]
\begin{center}
\begin{threeparttable}
\caption{Overview of datasets used in this study. The letters in the ``Usage'' column indicate the tasks for which the datasets were used. A: training auxiliary gender classifiers, B: SAN training, C: SAN evaluation, D: constructing unseen gender classifiers used for evaluating SAN models.}
\label{tab:datasets}
\centering
\small
\begin{tabular}{ l c c c}
 \toprule
 \multirow{2}{*}{ {\bf Dataset} } & {\bf \#male}& {\bf \#female}&   \multirow{2}{*}{ {\bf Usage} } \\
& {subjects / images} & {subjects / images} & \\
\midrule
CelebA-train & 4482 / \num[group-separator={,}]{73549} & 5163 / \num[group-separator={,}]{103772} & A, B\\ 
CelebA-test & 502 / \num[group-separator={,}]{7929} & 581 / \num[group-separator={,}]{11511}  & C\\ 
MORPH-train & \num[group-separator={,}]{10363} / \num[group-separator={,}]{41587} &  1938 / \num[group-separator={,}]{7567}  & D\\ 
MORPH-test & 1143 / \num[group-separator={,}]{4643} & 224 / 863 & C\\ 
LFW & 4205 / \num[group-separator={,}]{10064} & 1448 / \num[group-separator={,}]{2905}  & D\\ 
MUCT & 131 / \num[group-separator={,}]{1844} & 145 / \num[group-separator={,}]{1910}   & C\\ 
RaFD & 42 / \num[group-separator={,}]{1008} & 25 / 600 & C\\ 
\bottomrule 
\end{tabular} 
\end{threeparttable}
\end{center}
\end{table}

\subsection{Obtaining Race Labels}

Since race labels are not provided in the face datasets considered in this study, we designed a procedure to efficiently label the face images:
\begin{enumerate}[noitemsep]
\item Predict the racial labels for individual face images using a \textit{commercial-off-the-shelf} (COTS) software.
\item Aggregate the COTS predictions for each subject (for whom multiple face images with different poses are present in a given dataset) by majority voting. For example, if five face images of a given subject exist and three face images are labeled as \textit{white} and two face images are labeled as \textit{black}, the label \textit{white} was assigned to all five face images of the given subject.
\item Group the subjects based on their predicted majority class label from the previous step. Then, visually inspect one face image per subject and correct the class labels for all face images of a given subject if the class label was assigned incorrectly.
\end{enumerate}

\section{Experiments and Results}

As described in Section~\ref{sec:ensemble-san-formulation}, we trained and evaluated three auxiliary gender classifiers associated with the three ensemble SAN models: E1, E2, and E3. 
Table~\ref{tab:ensemble-aux-classifiers} summarizes the performance of these three models in terms of their gender classification errors on the CelebA-test and MORPH datasets. While the performance of E1 and E2 are similar, E3 outperforms E1 and E2 on MORPH. Given that $77\%$ of the face images in the MORPH dataset have the class label {\em black}, it is evident that oversampling examples of the under-represented race during training could have helped overcome the algorithmic bias in gender classification. 

\begin{figure*}
\begin{center}
   \includegraphics[width=0.82\linewidth]{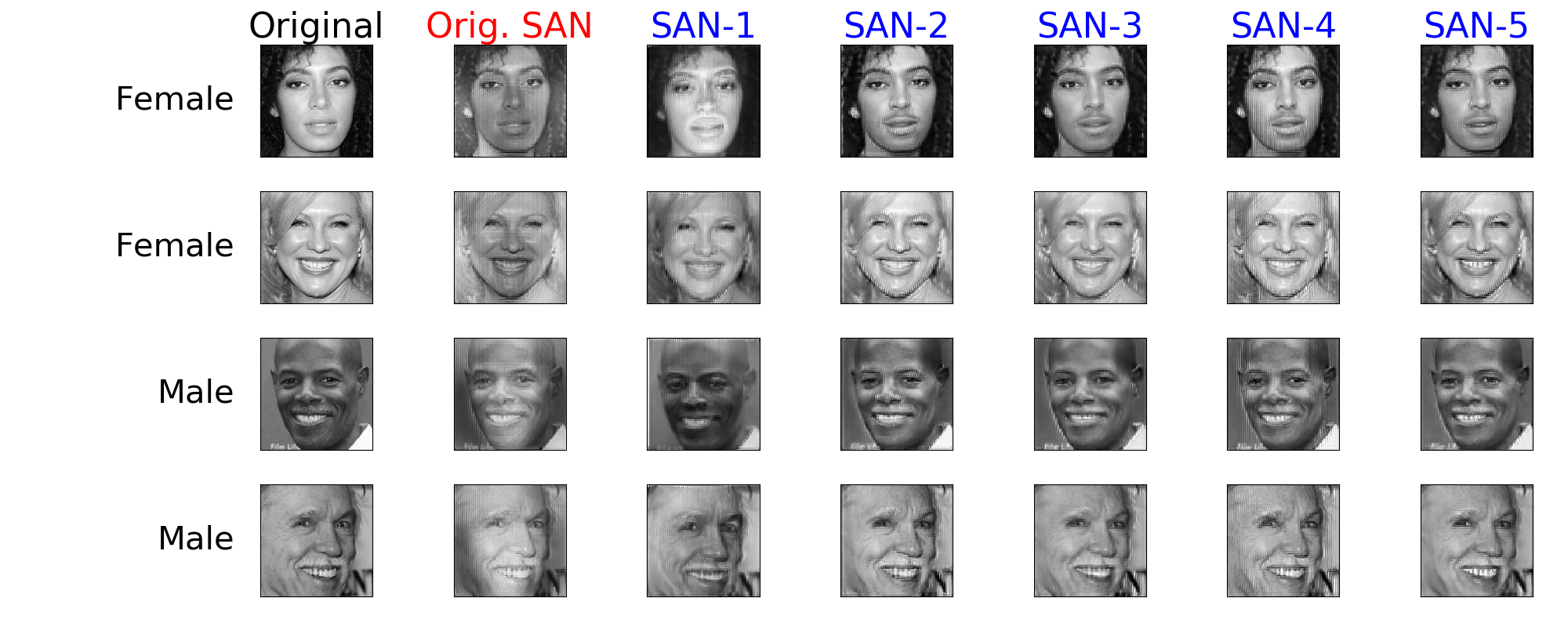}
\end{center}
   \caption{Four example images with their perturbed outputs using the original SAN model from Ref.~\cite{mirjalili_semi_2018} 
  and the outputs of five individual SAN models. Note that the ensemble SAN generates diverse outputs that is necessary for generalizing to arbitrary gender classifiers. }
\label{fig:example-orig-ensans}
\end{figure*}

Based on the results from Table~\ref{tab:ensemble-aux-classifiers}, the ensemble of auxiliary gender classifiers in E3 achieves higher accuracy on the MORPH-test dataset. In addition, in Table~\ref{tab:ensemble-aux-classifiers}, we  computed the  entropy as an empirical measure of ensemble diversity~\cite{kuncheva_combining_2004}, and the results confirm that auxiliary gender classifiers in E3 have higher diversity. Hence, we selected the ensemble SAN E3 for evaluation on unseen gender classifiers and face matchers. Figure~\ref{fig:example-orig-ensans} shows example images with their perturbed outputs from each of the SAN models in E3. In the remainder of the document, SAN-1 to SAN-5 denote the 5 models pertaining to E3.

\begin{table}
\caption{Error rates of the auxiliary gender classifiers on the CelebA / MORPH-test datasets. E3 (95\% confidence interval: 5.46\%--5.63\%) performs significantly better ($p \ll 0.01$) on the MORPH dataset compared to E1 (CI95: 6.24\%--6.42\%) and E2 (CI95: 6.25\%--6.43\%). At the end, ensemble diversities are reported~\cite{kuncheva_combining_2004}.}
\label{tab:ensemble-aux-classifiers}
\begin{center}
\begin{threeparttable}
\small
\begin{tabular}{cccc}
 \toprule
{\small Auxiliary} & {\bf E1:}& {\bf E2:}& {\bf E3:}\\
{\small  Classifier} &{\bf \small Regular} & {\bf \small Subject-based} & {\bf \small Race-based} \\ 
\midrule
$G^1$ & 2.25 / 5.56 & 2.07 / 6.24 & 2.29 / 5.17 \\ 
$G^2$ & 2.11 / 6.20 & 2.03 / 6.45 & 1.97 / 5.28 \\ 
$G^3$ & 2.03 / 6.38 & 2.06 / 6.46 & 2.13 / 5.04 \\ 
$G^4$ & 2.21 / 6.97 & 2.03 / 5.85 & 1.99 / 6.96 \\ 
$G^5$ & 2.42 / 6.53 & 2.12 / 6.72 & 2.02 / 5.28 \\ 
\midrule
Average: &  2.20 / 6.33 & 2.06 / 6.34 & 2.08 / \textbf{5.55} \\ 
Diversity: & 0.047/ 0.079 & 0.044 / 0.076 & 0.045 / {\bf 0.083} \\ \bottomrule
\end{tabular}
\end{threeparttable}
\end{center}
\end{table}

\label{sec:unseen-gender-classifiers}
\subsection{Unseen Gender Classifiers}
\begin{figure}
\begin{center}
   \includegraphics[width=0.95\linewidth]{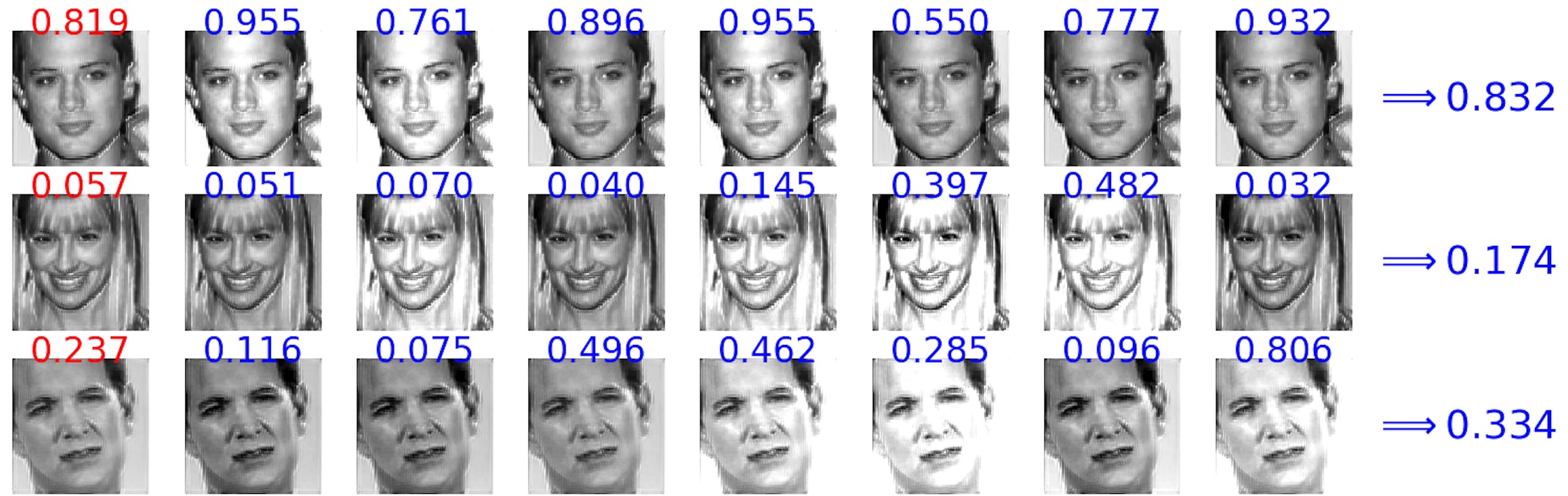}
\end{center}
   \caption{Data augmentation at the evaluation phase using random illumination and contrast adjustments. The left column shows the perturbed images before augmentation, and the next seven columns show the samples used for augmentation along with their gender prediction scores. Finally, average prediction scores obtained using the CNN-Merged model on these seven augmented samples are computed and denoted as CNN-Aug-Eval in the text.}
\label{fig:example-evalaug}
\end{figure}

In order to assess the performance of the proposed ensemble SAN in confounding an arbitrary gender classifier, we used 9 gender classifiers that were not available to any of the SAN models during training, as noted in Table~\ref{tab:unseen-gpreds}. We used five pre-trained models: a commercial-of-the-shelf gender classifier (G-COTS), IntraFace~\cite{de_la_torre_intraface_2015_long}, AFFACT~\cite{gunther_affact_2016}, and two additional Convolutional Neural Network(CNN)-based gender classifiers from Ref.~\cite{arriaga_real_2017}. In addition to the five existing gender classifiers, we also included CNN-based gender classifiers that were trained on three datasets, MORPH-train, LFW, and a merged version of MORPH-train and LFW. The CNN architecture of each of these gender classifiers contain five convolutional layers, each followed by SELU~\cite{klambauer_self_2017} activation units and a max-pooling layer. Inspired by HyperFace~\cite{ranjan_hyperface_2017}, the feature maps from the third convolution layer are fused with those of the last convolution layer to provide  features with hierarchical receptive fields for classification. The fused feature maps then undergo a global average pooling prior to two fully-connected layers, which were followed by a final sigmoid activation function. Two CNN models, named CNN-LFW and CNN-MORPH, were trained on the MORPH-train and LFW datasets, respectively, after the datasets were balanced by oversampling the female samples. A third CNN model, called CNN-Merged, was trained on the merged MORPH-train/LFW dataset, after balancing the male/female ratio, as well as balancing the size of the two datasets since MORPH-train is almost five times larger than LFW. Furthermore, we  also applied data augmentation during training by randomly adjusting illumination and contrast using the Torchvision library and PyTorch software~\cite{paszke_automatic_2017}. Finally, for the fourth gender predictor, we used CNN-Merged but applied data augmentation in the evaluation phase as suggested in~\cite{gunther_affact_2016}. Some examples of this data augmentation during evaluation are shown in Figure~\ref{fig:example-evalaug}. The illumination and contrast of a test sample is varied randomly to obtain seven samples. The augmented face images were then evaluated by the gender predictor, CNN-Merged, and the average score of the seven different modified test samples is reported; this is denoted as CNN-Aug-Eval (examples of the seven augmentation methods are shown in Figure~\ref{fig:example-evalaug}, columns 2-8).
\begin{table}
\begin{center}
\caption{List of the nine unseen gender classifiers used for evaluating the outputs of the proposed ensemble SAN models.}
\label{tab:unseen-gpreds}
\centering
\begin{tabular}{l|r}
\toprule
Pre-trained models & In-house trained CNN models\\
\midrule
G-COTS & CNN-MORPH\\
IntraFace~\cite{de_la_torre_intraface_2015_long} & CNN-LFW\\
AFFACT~\cite{gunther_affact_2016} & CNN-Merged\\
Ref.~\cite{arriaga_real_2017}-A & CNN-Aug-Eval\\
Ref.~\cite{arriaga_real_2017}-B & \\
\bottomrule
\end{tabular}
\end{center}
\end{table}

\begin{figure*}
\begin{center}
   \includegraphics[width=0.69\linewidth]{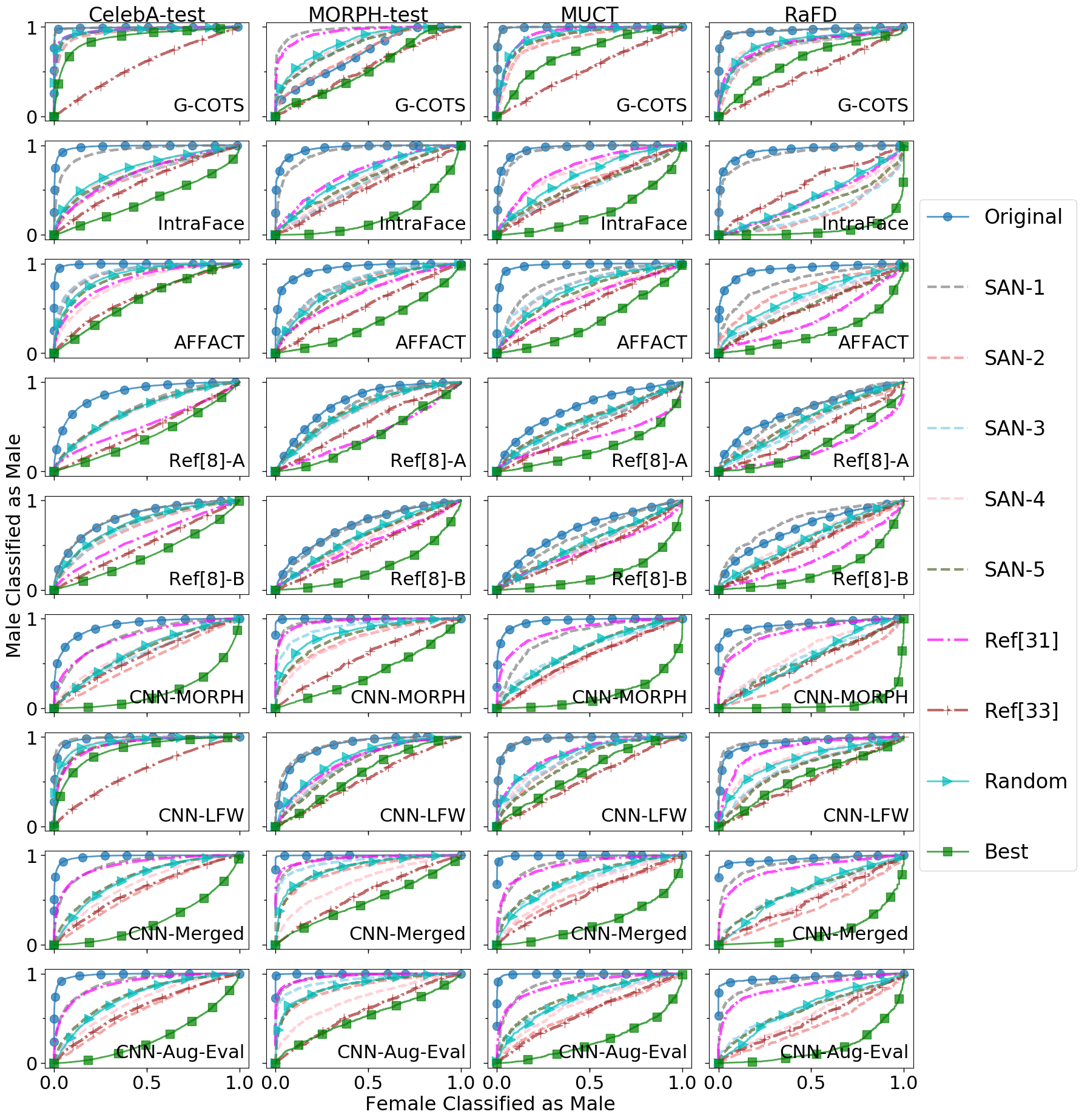}
\end{center}
   \caption{ROC curves of the nine unseen gender classifiers (each row corresponds to one classifier) on the perturbed images generated by each SAN model of the E3 ensemble on four evaluation datasets: CelebA-test, MORPH-test, MUCT, and LFW. Note that the gender classification performance shows a wide degree of change on perturbed samples, but in all cases, there is always one output from each ensemble degrading the performance.  }
\label{fig:roc-unseen-gpreds}
\end{figure*}

The performance of all nine unseen gender classifiers is shown in Figure~\ref{fig:roc-unseen-gpreds}. The ROC curves of gender prediction on the perturbed images generated by each SAN model is compared with the ROC curves of gender prediction on the original samples from the CelebA-test, MORPH-test, MUCT, and RaFD datasets. The ROC curves indicate that the gender classification performance varies widely across the SAN models. In certain cases, the perturbations made by some of the SAN models improve the performance of the gender classifier compared to their performance on the original data. In contrast to the original SAN model~\cite{mirjalili_semi_2018} (also shown in Figure~\ref{fig:roc-unseen-gpreds} for comparison), it is always possible to find at least one SAN model in the ensemble that can effectively degrade the gender classification performance for a given image. 

\begin{figure*}
\begin{center}
\includegraphics[width=0.68\linewidth]{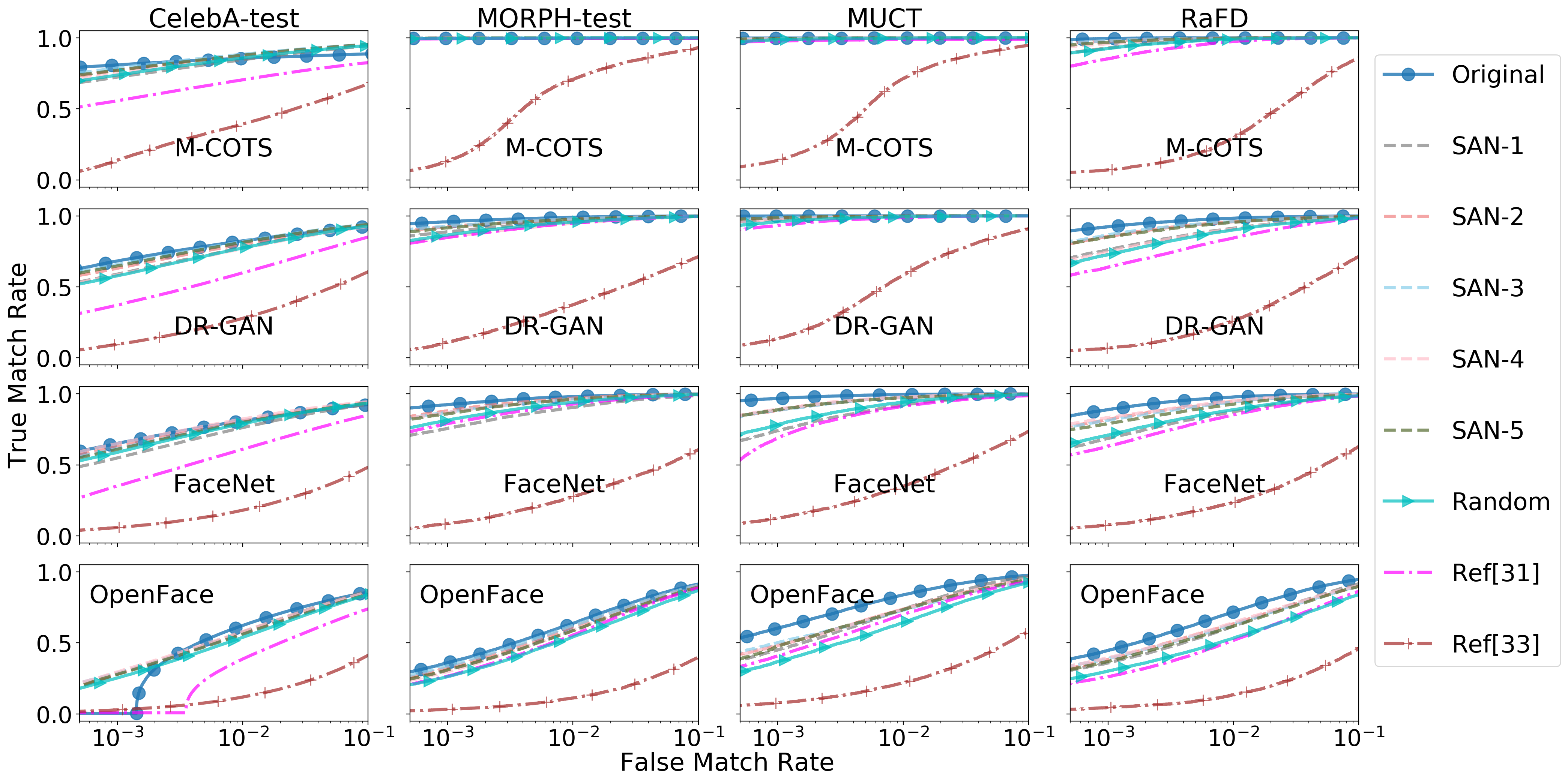}
\end{center}
   \caption{ROC curves of the four unseen face matchers (each row corresponds to one matcher) on the perturbed images generated by each SAN model of the E3 ensemble on four evaluation datasets: CelebA-test, MORPH-test, MUCT, and RaFD. Note that the matching performance is mostly retained except for some small degradations in the case of FaceNet and OpenFace. }
\label{fig:roc-unseen-matchers}
\end{figure*}

To illustrate the advantage of the proposed ensemble SAN over a single SAN, we did the following. For each unseen gender classifier, we selected the best-perturbed sample for each face image, $X$, based on the ground-truth gender label as follows:
\begin{equation}
P_{best} = \left\{
\begin{array}{ll} 
\displaystyle \min_{i=1..5}P(S^i(X)), & \text{if }
X \text{ is male;}\\
\displaystyle \max_{i=1..5}P(S^i(X)), & \text{if }
X \text{ is female.}
\end{array}\right.
\end{equation}

The ROC curve using the best-perturbed sample is shown in Figure~\ref{fig:roc-unseen-gpreds} for each gender classifier. The results suggests that diversity among individual SAN models is necessary for confounding unseen gender classifiers. 

\subsection{Unseen Face Matchers}

Next, we show the performance of unseen face matchers on the original and perturbed samples. For this analysis, we utilized four face matchers: a commercial-of-the-shelf face matcher (M-COTS) that has shown state-of-the-art performance in face recognition, and  face representation vectors obtained from DR-GAN~\cite{tran_disentangled_2017},  FaceNet~\cite{schroff_facenet_2015}, and OpenFace~\cite{amos_openface_2016}. For the latter three choices, we used the cosine-similarity measure between a pair of face vectors to measure their degree of dissimilarity.
Figure~\ref{fig:roc-unseen-matchers} shows the performance of these four matchers on the four evaluation datasets. The performance of M-COTS and DR-GAN on perturbed samples matches closely with that of original samples, except for some minor deviations for the DR-GAN matcher on the RaFD dataset. Performance of FaceNet and OpenFace on perturbed samples shows marginal deviation from that of original samples. In contrast, the face mixing approach~\cite{othman_privacy_2014} results in significant drop in performance of unseen face matchers, thereby suggesting that these outputs have lost their biometric utility.

{\bf Practical Implementation:} In a practical application, we may not have \emph{a priori} knowledge about the arbitrary gender classifier. Given an arbitrary gender classifier, one way to utilize the ensemble SAN is by randomly selecting one of the $t$ perturbed images. The result of such a random SAN model selection is shown in Figure~\ref{fig:roc-unseen-gpreds}. As the results illustrate, in most cases, randomly selecting a SAN model from the E3 ensemble results in better performance in terms of confounding arbitrary unseen gender classifiers compared to using a single SAN model. Furthermore, randomly selecting one SAN output does not degrade the face matching performance (Figure~\ref{fig:roc-unseen-matchers}).

Randomly selecting a perturbed sample tends to conceal the true gender label, since flipping the predicted label may or may not result in the true label of the original sample.

While gender recognition from a human perspective was not the main focus of this study, we may consider a human observer as an arbitrary gender classifier. The degree to which the gender information is concealed from human observers will be a subject of future studies.

\section{Conclusions}

In this work, we focused on addressing one of the main limitations of previous gender privacy methods, viz., their inability to generalize across multiple previously unseen gender classifiers. In this regard, we proposed an ensemble technique that generates diverse perturbations for an input face image, and at least one of the perturbed outputs is expected to confound the gender information with respect to an arbitrary gender classifier. We showed that randomly selecting perturbations for face images stored in a biometric database is an effective way for enabling gender privacy. In addition, we have showed that the face matching accuracy is retained for all perturbed outputs, thereby preserving the biometric utility of the face images.

For future work, we will extend the proposed privacy-preserving scheme to multiple soft-biometric attributes including age and race, and design a SAN model that can confound a selected combination of attributes while preserving matching performance.  This is expected to enhance the privacy of individuals whose biometric data is stored in central databases. 

\section{Acknowledgements}

This study is supported by the National Science Foundation under Grant Number $1618518$.

{\small
\bibliographystyle{ieee}
\balance
\bibliography{egbib}

\begin{thebibliography}{10}\itemsep=-1pt

\bibitem{eu_gdpr_2016}
{Regulation (EU) 2016/679 of the European Parliament and of the Council of 27
  April 2016}.
\newblock {\em {Official Journal of the European Union}}, L 119, 2016.

\bibitem{acquisti_privacy_2015}
A.~Acquisti, L.~Brandimarte, and G.~Loewenstein.
\newblock Privacy and human behavior in the age of information.
\newblock {\em Science}, 347(6221):509--514, 2015.

\bibitem{acquisti_predicting_2009}
A.~Acquisti and R.~Gross.
\newblock Predicting social security numbers from public data.
\newblock {\em Proceedings of the National Academy of Sciences},
  106(27):10975--10980, 2009.

\bibitem{acquisti_what_2013}
A.~Acquisti, L.~K. John, and G.~Loewenstein.
\newblock What is privacy worth?
\newblock {\em The Journal of Legal Studies}, 42(2):249--274, 2013.

\bibitem{acquisti_economics_2016}
A.~Acquisti, C.~Taylor, and L.~Wagman.
\newblock The economics of privacy.
\newblock {\em Journal of Economic Literature}, 54(2):442--92, 2016.

\bibitem{akhtar_threat_2018}
N.~Akhtar and A.~Mian.
\newblock Threat of adversarial attacks on deep learning in computer vision: A
  survey.
\newblock {\em arXiv preprint arXiv:1801.00553}, 2018.

\bibitem{amos_openface_2016}
B.~Amos, B.~Ludwiczuk, and M.~Satyanarayanan.
\newblock {OpenFace:} a general-purpose face recognition library with mobile
  applications.
\newblock {\em CMU School of Computer Science}, 2016.

\bibitem{arriaga_real_2017}
O.~Arriaga, M.~Valdenegro-Toro, and P.~Pl{\"o}ger.
\newblock Real-time convolutional neural networks for emotion and gender
  classification.
\newblock {\em arXiv preprint arXiv:1710.07557}, 2017.

\bibitem{becerra_age_2017}
F.~Becerra-Riera, H.~M{\'e}ndez-V{\'a}zquez, A.~Morales-Gonzalez, and
  M.~Tistarelli.
\newblock Age and gender classification using local appearance descriptors from
  facial components.
\newblock In {\em IEEE International Joint Conference on Biometrics}, pages
  799--804. IEEE, 2017.

\bibitem{bukar_automatic_2016}
A.~M. Bukar, H.~Ugail, and D.~Connah.
\newblock Automatic age and gender classification using supervised appearance
  model.
\newblock {\em Journal of Electronic Imaging}, 25(6):061605, 2016.

\bibitem{cao_partial_2017}
Z.~Cao, M.~Long, J.~Wang, and M.~I. Jordan.
\newblock Partial transfer learning with selective adversarial networks.
\newblock {\em arXiv preprint arXiv:1707.07901}, 2017.

\bibitem{castrillon_descriptors_2017}
M.~Castrill{\'o}n-Santana, J.~Lorenzo-Navarro, and E.~Ram{\'o}n-Balmaseda.
\newblock Descriptors and regions of interest fusion for in-and cross-database
  gender classification in the wild.
\newblock {\em Image and Vision Computing}, 57:15--24, 2017.

\bibitem{dantcheva_what_2016}
A.~Dantcheva, P.~Elia, and A.~Ross.
\newblock What else does your biometric data reveal? {A} survey on soft
  biometrics.
\newblock {\em {IEEE} Transactions on Information Forensics and Security},
  11(3):441--467, 2016.

\bibitem{de_la_torre_intraface_2015_long}
F.~De~la Torre, W.-S. Chu, X.~Xiong, F.~Vicente, X.~Ding, and J.~Cohn.
\newblock Intraface.
\newblock In {\em 11th {IEEE} International Conference on Automatic Face and
  Gesture Recognition ({FG})}, volume~1, pages 1--8, 2015.

\bibitem{dietterich_ensemble_2000}
T.~G. Dietterich.
\newblock Ensemble methods in machine learning.
\newblock In {\em International {W}orkshop on {M}ultiple {C}lassifier
  {S}ystems}, pages 1--15. Springer, 2000.

\bibitem{gunther_affact_2016}
M.~G{\"u}nther, A.~Rozsa, and T.~E. Boult.
\newblock {AFFACT}: Alignment free facial attribute classification technique.
\newblock {\em arXiv preprint arXiv:1611.06158}, 2016.

\bibitem{hansen_neural_1990}
L.~K. Hansen and P.~Salamon.
\newblock Neural network ensembles.
\newblock {\em IEEE {T}ransactions on {P}attern {A}nalysis and {M}achine
  {I}ntelligence}, 12(10):993--1001, 1990.

\bibitem{huang_labeled_2007_long}
G.~Huang, M.~Ramesh, T.~Berg, and E.~Learned-Miller.
\newblock Labeled faces in the wild: A database for studying face recognition
  in unconstrained environments.
\newblock Technical Report 07-49, University of Massachusetts, Amherst, October
  2007.

\bibitem{jain_introduction_2011}
A.~Jain, A.~Ross, and K.~Nandakumar.
\newblock {\em Introduction to biometrics}.
\newblock Springer Science \& Business Media, 2011.

\bibitem{jia_gender_2016}
S.~Jia, T.~Lansdall-Welfare, and N.~Cristianini.
\newblock Gender classification by deep learning on millions of weakly labelled
  images.
\newblock In {\em IEEE 16th International Conference on Data Mining Workshops},
  pages 462--467. IEEE, 2016.

\bibitem{jourabloo_attribute_2015}
A.~Jourabloo, X.~Yin, and X.~Liu.
\newblock Attribute preserved face de-identification.
\newblock In {\em International Conference on Biometrics ({ICB})}, pages
  278--285, 2015.

\bibitem{kindt_privacy_2016}
E.~J. Kindt.
\newblock {\em Privacy and data protection issues of biometric applications}.
\newblock Springer, 2013.

\bibitem{klambauer_self_2017}
G.~Klambauer, T.~Unterthiner, A.~Mayr, and S.~Hochreiter.
\newblock Self-normalizing neural networks.
\newblock In {\em Advances in Neural Information Processing Systems}, pages
  972--981, 2017.

\bibitem{kuncheva_combining_2004}
L.~I. Kuncheva.
\newblock {\em Combining pattern classifiers: methods and algorithms}.
\newblock John Wiley \& Sons, 2004.

\bibitem{langner_presentation_2010}
O.~Langner, R.~Dotsch, G.~Bijlstra, D.~H. Wigboldus, S.~T. Hawk, and
  A.~Van~Knippenberg.
\newblock Presentation and validation of the {R}adboud faces database.
\newblock {\em Cognition and emotion}, 24(8):1377--1388, 2010.

\bibitem{levi_age_2015}
G.~Levi and T.~Hassner.
\newblock Age and gender classification using convolutional neural networks.
\newblock In {\em Proceedings of the {IEEE} Conference on Computer Vision and
  Pattern Recognition Workshops}, pages 34--42, 2015.

\bibitem{facebook_scandal_2018}
P.~Lewis and J.~Carrie~Wong.
\newblock {Facebook employs psychologist whose firm sold data to Cambridge
  Analytica}.
\newblock
  {https://www.theguardian.com/news/2018/mar/18/facebook-cambridge-analytica-joseph-chancellor-gsr},
  2018.
\newblock [Online; accessed 26-April-2018].

\bibitem{liu_deep_2015_long}
Z.~Liu, P.~Luo, X.~Wang, and X.~Tang.
\newblock Deep learning face attributes in the wild.
\newblock In {\em Proceedings of the {IEEE} International Conference on
  Computer Vision}, pages 3730--3738, 2015.

\bibitem{mansanet_local_2016}
J.~Mansanet, A.~Albiol, and R.~Paredes.
\newblock Local deep neural networks for gender recognition.
\newblock {\em Pattern Recognition Letters}, 70:80--86, 2016.

\bibitem{milborrow_muct_2010}
S.~Milborrow, J.~Morkel, and F.~Nicolls.
\newblock The {MUCT} landmarked face database.
\newblock {\em {PRASA}}, 2010.

\bibitem{mirjalili_semi_2018}
V.~Mirjalili, S.~Raschka, A.~Namboodiri, and A.~Ross.
\newblock Semi-{A}dversarial {N}etworks: {C}onvolutional autoencoders for
  imparting privacy to face images.
\newblock In {\em Proc. of 11th IAPR International Conference on Biometrics
  (ICB)}, Gold Coast, Australia, 2018.

\bibitem{mirjalili_soft_2017}
V.~Mirjalili and A.~Ross.
\newblock Soft biometric privacy: Retaining biometric utility of face images
  while perturbing gender.
\newblock In {\em Proc. of International Joint Conference on Biometrics
  (IJCB)}, 2017.

\bibitem{othman_privacy_2014}
A.~Othman and A.~Ross.
\newblock Privacy of facial soft biometrics: Suppressing gender but retaining
  identity.
\newblock In {\em European Conference on Computer Vision Workshop}, pages
  682--696. Springer, 2014.

\bibitem{parkhi_deep_2015}
O.~M. Parkhi, A.~Vedaldi, and A.~Zisserman.
\newblock Deep face recognition.
\newblock In {\em British Machine Vision Conference}, volume~1, page~6, 2015.

\bibitem{paszke_automatic_2017}
A.~Paszke, S.~Gross, S.~Chintala, G.~Chanan, E.~Yang, Z.~DeVito, Z.~Lin,
  A.~Desmaison, L.~Antiga, and A.~Lerer.
\newblock Automatic differentiation in pytorch.
\newblock In {\em NIPS-W}, 2017.

\bibitem{ranjan_hyperface_2017}
R.~Ranjan, V.~M. Patel, and R.~Chellappa.
\newblock {HyperFace}: A deep multi-task learning framework for face detection,
  landmark localization, pose estimation, and gender recognition.
\newblock {\em IEEE Transactions on Pattern Analysis and Machine Intelligence},
  2017.

\bibitem{raschka_python_2017}
S.~Raschka and V.~Mirjalili.
\newblock {\em {Python machine learning, 2nd Ed.}}
\newblock Packt Publishing, Birmingham, UK, 2017.

\bibitem{ricanek_morph_2006}
K.~Ricanek and T.~Tesafaye.
\newblock {MORPH}: A longitudinal image database of normal adult
  age-progression.
\newblock In {\em 7th International Conference on Automatic Face and Gesture
  Recognition}, pages 341--345, 2016.

\bibitem{rozsa_are_2016_long}
A.~Rozsa, M.~G\"unther, E.~M. Rudd, and T.~E. Boult.
\newblock Are facial attributes adversarially robust?
\newblock {\em {arXiv} preprint {arXiv}:1605.05411}, 2016.

\bibitem{schroff_facenet_2015}
F.~Schroff, D.~Kalenichenko, and J.~Philbin.
\newblock {FaceNet:} a unified embedding for face recognition and clustering.
\newblock In {\em Proceedings of the IEEE {C}onference on {C}omputer {V}ision
  and {P}attern {R}ecognition}, pages 815--823, 2015.

\bibitem{sim_controllable_2015}
T.~Sim and L.~Zhang.
\newblock Controllable face privacy.
\newblock In {\em 11th {IEEE} International Conference on Automatic Face and
  Gesture Recognition ({FG})}, volume~4, pages 1--8, 2015.

\bibitem{strauss_ensemble_2017}
T.~Strauss, M.~Hanselmann, A.~Junginger, and H.~Ulmer.
\newblock Ensemble methods as a defense to adversarial perturbations against
  deep neural networks.
\newblock {\em arXiv preprint arXiv:1709.03423}, 2017.

\bibitem{szegedy_intriguing_2013}
C.~Szegedy, W.~Zaremba, I.~Sutskever, J.~Bruna, D.~Erhan, I.~Goodfellow, and
  R.~Fergus.
\newblock Intriguing properties of neural networks.
\newblock {\em {arXiv} preprint {arXiv}:1312.6199}, 2013.

\bibitem{tran_disentangled_2017}
L.~Tran, X.~Yin, and X.~Liu.
\newblock Disentangled representation learning {GAN} for pose-invariant face
  recognition.
\newblock In {\em Computer vision and pattern recognition ({CVPR})}, 2017.

\end{thebibliography}
}

\end{document}